\documentclass[runningheads]{llncs}
\usepackage[T1]{fontenc}
\usepackage{graphicx}
\usepackage{algorithm}
\usepackage[noend]{algorithmic}
\usepackage{amsmath}
\usepackage{amssymb}
\usepackage{caption}  
\usepackage{hyperref}
\usepackage{xcolor}

\usepackage{tikz}  
\usepackage{textpos} 
\usepackage{color}

\begin{document}
\title{Understanding Federated Learning from IID to Non-IID dataset: An Experimental Study}
%
%
\author{Jungwon Seo, Ferhat Ozgur Catak \and
Chunming Rong }
\authorrunning{Seo et al.}
%
\institute{University of Stavanger, Norway\\
\email{\{jungwon.seo, f.ozgur.catak, chunming.rong\}@uis.no}
}

\maketitle              
\begin{textblock*}{\textwidth}(-40mm,-70mm)  
    \footnotesize \textit{Presented at the 36th NIKT
25–27 November 2024, Bergen, Norway}
\end{textblock*}

\begin{abstract}
As privacy concerns and data regulations grow, federated learning (FL) has emerged as a promising approach for training machine learning models across decentralized data sources without sharing raw data. However, a significant challenge in FL is that client data are often non-IID (non-independent and identically distributed), leading to reduced performance compared to centralized learning. While many methods have been proposed to address this issue, their underlying mechanisms are often viewed from different perspectives. Through a comprehensive investigation from gradient descent to FL, and from IID to non-IID data settings, we find that inconsistencies in client loss landscapes primarily cause performance degradation in non-IID scenarios. From this understanding, we observe that existing methods can be grouped into two main strategies: (i) adjusting parameter update paths and (ii) modifying client loss landscapes. These findings offer a clear perspective on addressing non-IID challenges in FL and help guide future research in the field.

\end{abstract}

\section{Introduction}

Federated Learning (FL) is a distributed machine learning framework that prioritizes privacy preservation~\cite{mcmahan2017communication}. Unlike traditional centralized methods, where client data is aggregated on a central server for model training, FL allows clients to keep their data locally. Clients collaborate by sharing only their locally trained models with a central server, which aggregates these models to build a global model. This approach enables learning from diverse, distributed datasets without exposing data, making FL especially valuable for privacy-sensitive applications.

Although similar distributed learning frameworks have existed before~\cite{chu2006map,zinkevich2010parallelized,dean2012large}, the term "Federated Learning" was officially coined by McMahan et al. in 2016~\cite{mcmahan2017communication}. Since then, particularly in deep learning, FL has gained significant attention in both research and industry~\cite{kairouz2021advances}. One of the primary challenges in FL is the heterogeneity of data across clients~\cite{zhao2018federated}. When client data distributions are non-IID (not Independent and Identically Distributed), the learning direction of individual clients may diverge from that of the global model. This divergence, often referred to as "\textit{client drift}", can degrade the overall performance of the global model and, in extreme cases, may even hinder convergence~\cite{karimireddy2020scaffold}. Numerous studies have addressed this issue, primarily from an optimization perspective, proposing both theoretical and empirical methods~\cite{li2020federated,karimireddy2020scaffold,acar2021federated}.

As the field has matured, a variety of new perspectives have emerged. Some researchers argue that FL’s challenges resemble those found in catastrophic forgetting, similar to continual learning~\cite{lee2022preservation,lee2024fedsol}, while others suggest that the smoothness of the loss function, inspired by sharpness-aware training, plays a crucial role~\cite{qu2022generalized,sun2023dynamic}. Additionally, some point to differences in learned representations, relating to concepts from contrastive learning~\cite{li2021model}.

Initially, the FL problem was perceived as a straightforward optimization task. However, the incorporation of new perspectives has introduced additional layers of complexity, making it challenging to identify the critical areas of focus. Moreover, most existing studies primarily investigate FL under non-IID settings, leaving a gap in our understanding of how FL behaves from scratch. As a result, FL has become a "black box within a black box," exacerbating the opacity already present in deep learning models. This multitude of perspectives has made it difficult for researchers to interpret proposed methods and to determine the most promising research directions.

To address this, we propose a systematic investigation of FL, starting with IID settings and gradually exploring its behavior across various hyperparameters and conditions. By grounding our analysis in gradient descent and extending it to FL, we aim to uncover the mechanisms driving its performance and explain the unexpected phenomena observed in complex, non-IID environments. Through this approach, we hope to clarify FL’s core challenges and offer direction for future research.

The contributions of this study are as follows:
\vspace{-2mm}
\begin{itemize}
\item \textbf{Optimization Algorithm Review}: We provide a clear review of optimization algorithms, from gradient descent to FedAvg.

\item \textbf{Experiments in IID Settings}: We conduct extensive experiments on FedAvg under IID conditions, exploring various hyperparameters and FL settings to gain insights into its behavior.

\item \textbf{Non-IID Analysis}: Using insights from the IID experiments, we analyze FL performance in non-IID settings, interpreting the observed phenomena.

\item \textbf{Categorizing Approaches for Non-IID Challenges}: We classify existing and future methods into two strategies: (a) update path adjustment and (b) loss landscape modification, addressing non-IID data in FL.
\end{itemize}
\vspace{-3mm}

\section{From Gradient Descent to FedAvg}

Gradient Descent (GD) is one of the most fundamental algorithms in machine learning for optimizing model parameters, such as those in neural networks~\cite{ruder2016overview}. Consider an objective function $F(\theta)$ that we seek to minimize, where $\theta$ represents the model parameters, with $\theta \in \mathbb{R}^d$, and $d$ is the dimensionality of $\theta$. The parameter update rule in GD is given by:

\begin{equation}\label{eq:gd}
    \theta^{t+1} = \theta^{t} - \eta \nabla F(\theta^t; \mathcal{D})
\end{equation}

Here, $t$ denotes the iteration step, and $\nabla F(\theta^t; \mathcal{D})$ represents the gradient of the objective function with respect to $\theta^t$, evaluated on the dataset $\mathcal{D}$. By updating the parameters in the negative direction of the gradient with a learning rate $\eta$, the parameters move towards a more optimal solution. This iterative process is repeated until the gradient $\nabla F(\theta^t)$ converges to zero, indicating that the model has reached a local or global minimum.

A major limitation of GD lies in its dependence on a static dataset $\mathcal{D}$. This becomes problematic when the dataset is too large to fit into a single device's memory or when full-batch training becomes computationally expensive. GD requires computing the gradient over all data points and using the averaged gradient for each update at iteration $t$. Expanding the update rule in Eq.~\ref{eq:gd}, we obtain:

\begin{equation}\label{eq:exp-gd} \theta^{t+1} = \theta^{t} - \eta \frac{1}{|\mathcal{D}|} \sum_{i \in \mathcal{D}} \nabla F(\theta^t; x_i,y_i) \end{equation}

Here, $x_i$ and $y_i$ represent the feature and label pairs from the dataset $\mathcal{D}$ in a supervised learning scenario. Eq.~\ref{eq:exp-gd} highlights that, before each update, the gradient must be computed and averaged across all data points. Each data point contributes a distinct gradient, and the average of these gradients determines the final update direction. This gradient averaging is fundamental to optimization and closely relates to the parameter averaging concept in Federated Averaging (FedAvg). FedAvg can thus be viewed as a generalization of traditional optimization methods, a connection that will be elaborated upon.

To address GD's inefficiencies, SGD was introduced. Unlike GD, which updates after computing the gradient over the entire dataset, SGD updates the parameters after each individual data sample. This leads to faster updates and reduced memory usage, making it suitable for large-scale problems. However, this method can be unstable, as computing gradients from a single data point may introduce bias and slow convergence in terms of model performance, despite the higher computational efficiency.

\begin{figure}
\centering
\includegraphics[width=0.8\textwidth]{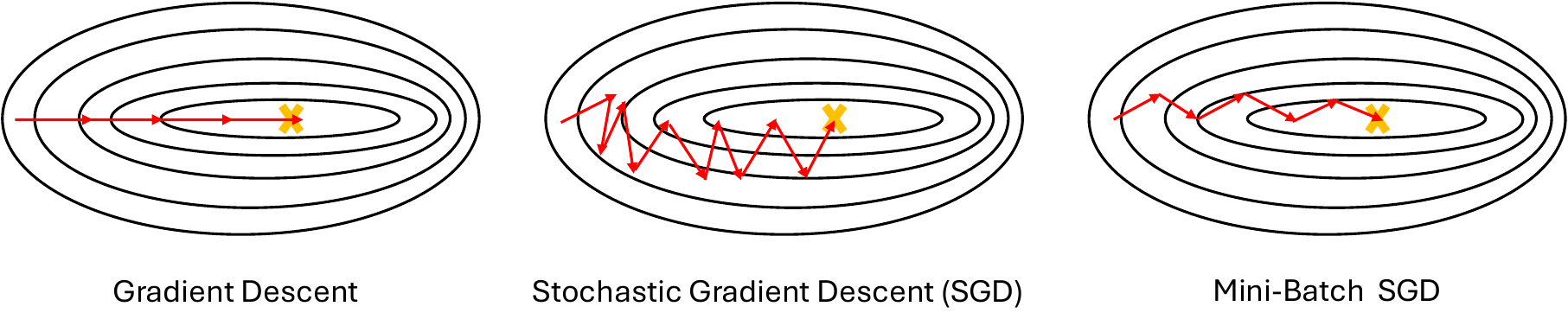}

\caption{Illustration of how different GD methods navigate the loss landscape} \label{fig:gd-to-msgd}
\end{figure}
\vspace{-3mm}

To balance the trade-offs between GD and SGD, mini-batch SGD was developed. It computes gradients using a batch of size $B$ for each update, reducing SGD's instability while improving GD's efficiency. GD, SGD, and mini-batch SGD can be seen as special cases within a broader framework, where the batch size $B$ ranges from 1 (SGD) to the entire dataset (GD), as shown in Algorithm \ref{alg.sgd}, with their respective optimization paths illustrated in Fig.~\ref{fig:gd-to-msgd}.

\begin{algorithm}\label{alg:sgd}

\caption{Stochastic Gradient Descent (SGD)}\label{alg.sgd}
\begin{algorithmic}[1]
\STATE \textbf{Input:} Learning rate $\eta$, Batch size $B$, Number of epochs $E$
\FOR{$n = 1$ to $E$}
\FOR{each random mini-batch $\mathcal{B}$ of size $B$ with data $(x_i, y_i)$}
\STATE Compute gradient $g = \frac{1}{B} \sum_{i \in \mathcal{B}} \nabla_\theta F(\theta; x_i, y_i)$
\STATE Update parameters: $\theta \leftarrow \theta - \eta \cdot g$
\ENDFOR
\ENDFOR
\end{algorithmic}
\end{algorithm}


The concept of gradient averaging for training extends naturally to distributed and parallel computing, as gradient calculations are independent until aggregation. Scaling the SGD process across multiple devices (e.g., GPUs or distributed systems) accelerates training and mitigates the limitations of centralized hardware. While there are various approaches to parallel SGD~\cite{chu2006map,zinkevich2010parallelized,dean2012large}, in this work, we focus on aggregating gradients from each device per batch of each client, as described in Algorithm~\ref{alg:parallelsgd}. In this distributed learning approach, we introduce a parameter, $K$, representing the number of devices. Gradients are computed on each device, followed by averaging locally and then globally.

\vspace{-3mm}
\begin{algorithm}
\caption{Parallel Stochastic Gradient Descent (Parallel SGD) 
}
\begin{algorithmic}[1]\label{alg:parallelsgd}

\STATE \textbf{Input:} $\eta$, $B$, $E$, Number of devices $K$
\FOR{$n = 1$ to $E$}
    \FOR{each device $k = 1$ to $K$ \textbf{in parallel}}
        \STATE Uniformly select a random mini-batch $B_k$ from $\mathcal{D}_k$
        
        \STATE Compute gradient on device $k$: $g_k = \frac{1}{|B_k|} \sum_{i \in B_k} \nabla_\theta F(\theta; x_i, y_i)$
    \ENDFOR
    \STATE Aggregate gradients: $g = \frac{1}{K} \sum_{k=1}^{K} g_k$
    \STATE Update parameters: $\theta \leftarrow \theta - \eta \cdot g$
\ENDFOR
\end{algorithmic}
\end{algorithm}
\vspace{-4mm}

Despite the potential benefits of distributed and parallel processing, communication bottlenecks can arise, particularly during gradient aggregation. While intra-device parallelism (e.g., multi-GPU operations) tends to be efficient, frequent gradient communication between devices introduces overhead, offsetting the advantages of parallelization~\cite{sergeev2018horovod}.

A natural approach to mitigate this issue is to reduce the communication frequency. Several studies have investigated methods where each device updates parameters independently for a set duration, merging gradients less frequently. In LocalSGD~\cite{stich2018local}, each device performs training for $I$ local epochs, followed by gradient merging every $I+1$ epochs. The globally updated model is then used as the starting point for subsequent training, as described in Algorithm~\ref{alg:localsgd}.

\begin{algorithm}
\caption{Local Stochastic Gradient Descent (Local SGD) }
\begin{algorithmic}[1]\label{alg:localsgd}
\STATE \textbf{Input:} $\eta$, $B$, $E$, $K$, Synchronization period $I$
\FOR{$n = 1$ to $E$}
    \FOR{each device $k = 1$ to $K$ \textbf{in parallel}}
        \STATE Initialize local parameters: $\theta_k \leftarrow \theta$
        \FOR{$t = 1$ to $I$}
            \STATE Uniformly select a random mini-batch $B_k^t$ from $\mathcal{D}_k$
            \STATE Compute gradient on device $k$: $g_k = \frac{1}{|B_k^t|} \sum_{i \in B_k^t} \nabla_\theta F(\theta_k; x_i, y_i)$
            \STATE Update local parameters: $\theta_k \leftarrow \theta_k - \eta \cdot g_k$
        \ENDFOR
    \ENDFOR
    \STATE Synchronize: Aggregate parameters across devices: $\theta \leftarrow \frac{1}{K} \sum_{k=1}^{K} \theta_k$
\ENDFOR
\end{algorithmic}
\end{algorithm}

LocalSGD can be viewed as a special case of FedAvg, more aligned with distributed learning. It uses centrally managed, evenly partitioned IID data across all devices, reducing gradient variance and closely approximating centralized training performance. While sharing FedAvg's core mechanics, LocalSGD assumes balanced IID data, making it more suited for distributed learning. 

In contrast, FedAvg (Algorithm.~\ref{alg:fedavg}) assumes partial participation of client (Line~\ref{line:partial}) and non-IID data across clients, requiring adjustments in gradient aggregation. Client updates are weighted by dataset size, with larger datasets contributing more to the global model, as noted in line~\ref{line:agg2}.

\vspace{-3mm}
\begin{algorithm}
\caption{Federated Averaging (FedAvg)}
\begin{algorithmic}[1]\label{alg:fedavg}
\STATE \textbf{Input:} Global $\eta_g$, Local $\eta_l$, $B$, $E$, $K$, rounds $R$, client sampling ratio $C$

\FOR{each round $r = 1$ to $R$}
    \STATE Randomly select a subset $\mathcal{S}$ of $\max(C \cdot K, 1)$ clients\label{line:partial}

    \FOR{each client $k \in \mathcal{S}$ \textbf{in parallel}}
        \STATE Initialize local model: $\theta_k \gets \theta$
        \FOR{each local epoch $n = 1$ to $E$}
            \FOR{each mini-batch $B_k$ from local dataset $\mathcal{D}_k$}
                \STATE Compute gradient: $g_k = \frac{1}{|B_k|} \sum_{(x_i, y_i) \in B_k} \nabla_\theta F(\theta_k; x_i, y_i)$
                \STATE Update local model: $\theta_k \gets \theta_k - \eta_l \cdot g_k$
            \ENDFOR
        \ENDFOR
        \STATE Send back to server: (Option I) $\Delta \theta_k = \theta_k - \theta$ or (Option II) $ \theta_k$
    \ENDFOR
    \STATE Aggregate the total number of data points: $M = \sum_{k \in \mathcal{S}} |\mathcal{D}_k|$
    \STATE Update global model: (I) $\theta \gets \theta + \eta_g \cdot \sum_{k \in \mathcal{S}} \frac{|\mathcal{D}_k|}{M} \cdot \Delta \theta_k$ or (II) $\theta \gets \sum_{k \in \mathcal{S}} \frac{|\mathcal{D}_k|}{M} \cdot \theta_k$ \label{line:agg2}
\ENDFOR
\end{algorithmic}
\end{algorithm}
\vspace{-3mm}
Originally, FedAvg exchanges local model parameters (Option II) instead of updates (Option I), which are mathematically equivalent. However, sending updates allows for more flexibility in modifying the aggregation process, enabling techniques such as global learning rate adjustments and optimizers~\cite{karimireddy2020scaffold,reddi2021adaptive}. Moreover, this formulation provides a clearer conceptual framework for understanding FL. From the perspective of Option I in line~\ref{line:agg2}, the updates from each client resemble gradients from individual data samples in GD. Therefore, FL is not merely parameter mixing but remains a structured optimization process. The following sections provide an empirical analysis of how FL works in practice.

\vspace{-3mm}
\section{Experimental Setup}

\subsection{Basic Configuration} 

We conduct our experiments using the CIFAR-10 dataset (50,000 instances in the training set and 10,000 in the test set), a standard benchmark in FL research. The model architecture is a convolutional neural network (CNN) with two convolutional layers followed by ReLU activations and max-pooling. It includes two fully connected layers, leading to a 10-class output, as described in \cite{mcmahan2017communication}. The task is a classification problem, with cross-entropy loss serving as the objective function. To ensure a controlled experimental environment, we deliberately exclude additional techniques such as weight decay, data augmentation, and advanced optimizers (e.g., momentum or adaptive techniques). This approach allows us to isolate the impact of the hyperparameters $B$, $E$, and $\eta$ on the SGD process.

\vspace{-3mm}
\subsection{FL Setup}

\subsubsection{Evaluation}\label{sec:eval}
We evaluate model performance by analyzing both convergence speed and generalization quality. While rapid convergence is desirable, it does not inherently guarantee strong generalization, making it crucial to track both aspects throughout training. In our setup, global model accuracy (Top-1) is measured server-side using a dedicated test set. Test loss is also computed server-side to assess generalization, while training loss is computed locally by clients. The server averages client-side training losses to monitor overall optimization progress.

\vspace{-3mm}

\subsubsection{Transition from Centralized to Federated Learning}\label{sec:exp-setup}

To set initial hyperparameters, we first establish reasonable baseline performance in a centralized learning (CL) setup, corresponding to $K=1$. We use $B=500$, $\eta=0.005$, and $E=1$, yielding a final accuracy of around 68\%. These hyperparameters are not fine-tuned for peak performance in CL; rather, the aim is to observe how performance evolves as we transition from CL to FL.

\begin{figure} \includegraphics[width=\linewidth]{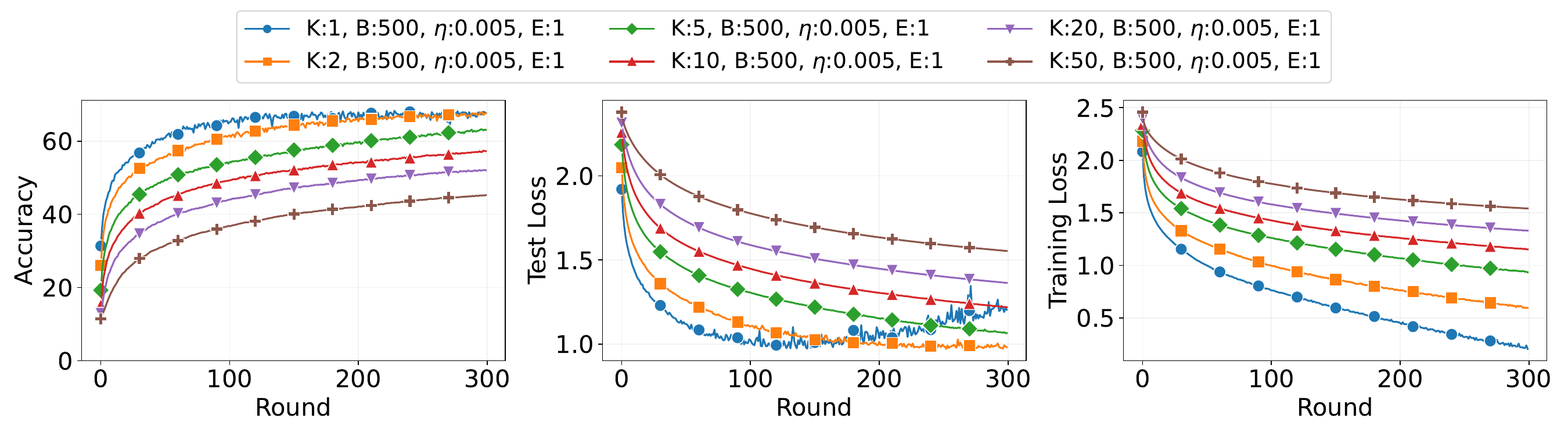} 
\vspace{-6mm}
\caption{Impact of varying $K$ with consistent hyperparameters.} 
\label{fig:cl-to-fl} 
\end{figure}

When $K$ changes, the dataset is proportionally divided among $K$ clients. For $K=1$ (CL), a single client holds the entire training dataset of 50,000 samples, while for $K=10$, each client is allocated 5,000 unique training samples. We initialize with a balanced IID dataset, ensuring that all clients have an equal distribution of labels and an identical number of samples. The effect of varying the number of clients on FL training dynamics is shown in Fig.~\ref{fig:cl-to-fl}.

\vspace{-4mm}
\vspace{-5mm}

In this setup, as $K$ increases, there is a noticeable decline in performance within a fixed number of communication rounds, highlighting a common oversight. When transitioning from CL to FL with a limited dataset, the critical factor that must be consistently controlled for a fair comparison is the \textbf{amount} of effective updates per round, $u$, rather than the \textbf{number}, due to the influence of $\eta$. It is defined as:

\begin{equation}\label{eq:local-update} 
u = \eta \cdot \frac{E \cdot |\mathcal{D}|}{B \cdot K} 
\end{equation}

Therefore, local epoch $E$, batch size $B$, and learning rate $\eta$ must be adjusted based on each client's dataset size $\frac{|\mathcal{D}|}{K}$ to ensure all scenarios achieve an equal $u$.

\begin{figure} \includegraphics[width=\linewidth]{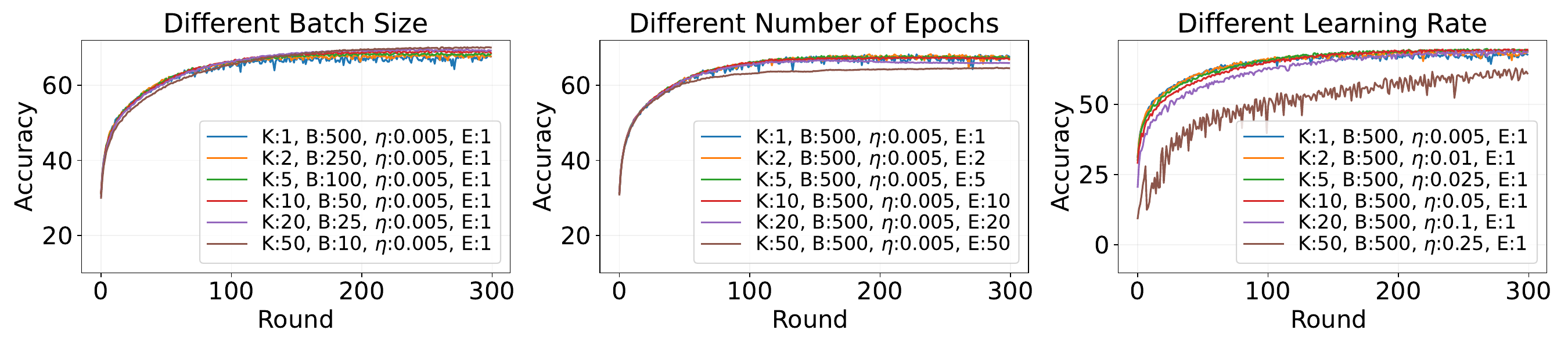} 
\vspace{-3mm}
\caption{Impact of matched $B$, $E$, and $\eta$ on performance with different $K$.} \label{fig:diff-param}
\end{figure}

In Fig.~\ref{fig:diff-param}, we present three accuracy charts corresponding to adjustments in $B$, $E$, and $\eta$. Unlike the previous fixed hyperparameter setting, these adjustments align performance and convergence rates across different client counts. When $E$ is increased to 50 and $\eta$ is set to 0.25, it results in an extreme case, which we disregard in our analysis. Unless specified otherwise, the base setting for all experiments is $K = 10$, $E = 1$, $B = 50$, and $\eta = 0.005$, with variations applied to a single parameter in each experiment.

\section{Experimental Results}
\subsection{Impact of Hyperparameters under IID Condition} 
\begin{figure}
\includegraphics[width=\linewidth]{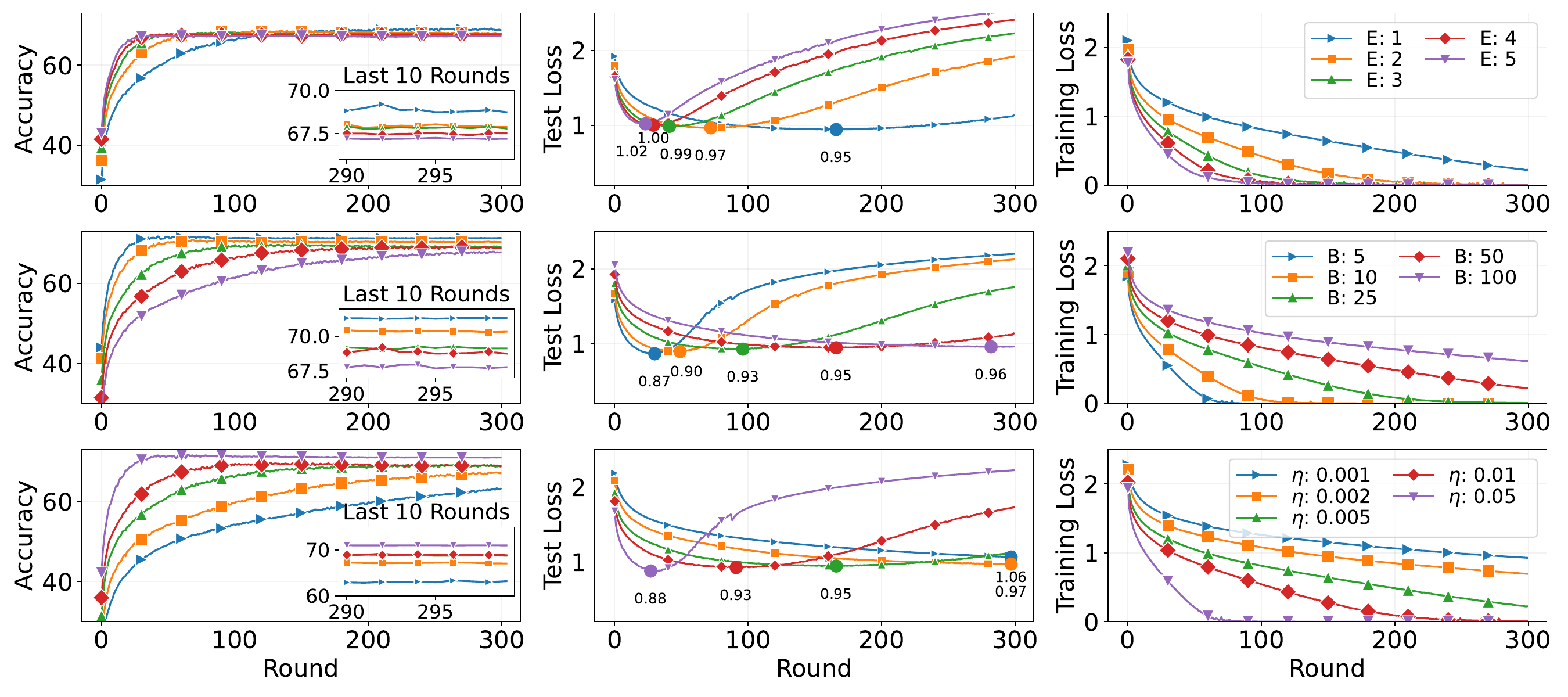}
\caption{Training dynamics when $K=10$. The first row shows results for varying $E$, the second for different $B$, and the third for various $\eta$.}

\label{fig:iid-diff-param}
\end{figure}

We first assess how three hyperparameters—$E$, $\eta$, and $B$—affect test accuracy, test loss, and training loss. As illustrated in Fig.~\ref{fig:iid-diff-param}, increasing $E$ or $\eta$, or decreasing $B$, speeds up convergence. This is evident from the curvature of the accuracy plots and the more rapid decline in training loss, which is consistent with our update amount equation $u$ (Eq.~\ref{eq:local-update}). Notably, the point of overfitting occurs earlier with faster convergence, as marked by the minimum test loss (indicated by the circle).

However, some notable observations can be made. While higher $E$ values result in faster convergence, they lead to suboptimal minimum test loss and final accuracy, with lower epochs (e.g., $E=1$) providing better performance (first row in Fig.~\ref{fig:iid-diff-param}). In contrast, the smallest batch size ($B=5$) achieves the highest final accuracy and the lowest test loss (second row in Fig.~\ref{fig:iid-diff-param}). Similarly, higher $\eta$ produce the best final accuracy and the lowest test loss (third row in Fig.~\ref{fig:iid-diff-param}) while both of them provide faster convergence. Although the differences are not substantial under IID conditions, they become more pronounced in non-IID settings, which we will discuss later.

\vspace{-3mm}
\subsection{Partial Participation and Imbalanced Data under IID Condition}
\vspace{-2mm}
This section explores the effects of partial client participation and imbalanced data in the context of FL under IID conditions. While these are not hyperparameters, they are key aspects of the experimental setup. In FL, the server cannot always ensure that all clients participate in every training round. As a result, the server either waits for all clients or proceeds with only those that respond in time, leading to partial participation (PP), where only a subset of clients contribute to model updates in each round.

Even under IID conditions, where each client has an equal proportion of labels, the number of data samples per client can vary, creating imbalanced datasets. Some clients may have significantly more data than others.

We now examine how partial participation and data imbalance influence training dynamics, particularly with regard to convergence and model performance.
\vspace{-4mm}
\subsubsection{Impact of Partial Participation}\label{sec:iid-pp}

\begin{figure} \includegraphics[width=\linewidth]{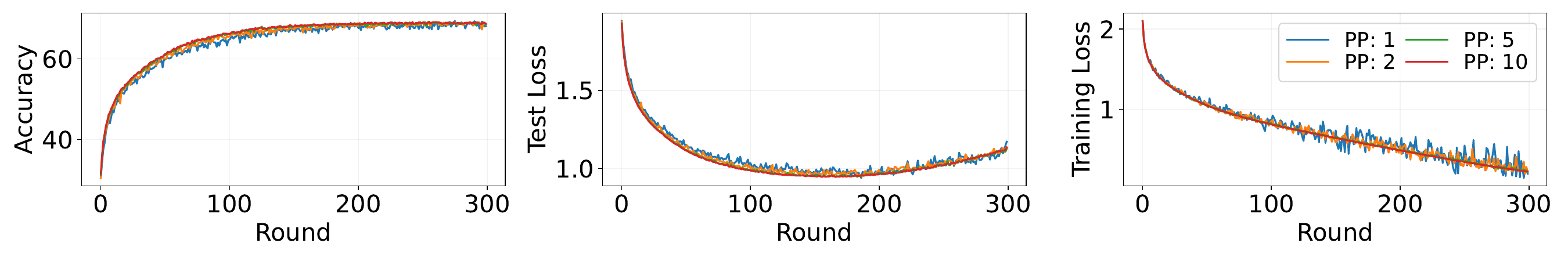} 
\caption{Training Dynamics under Partial Participation with IID Condition} \label{fig:pp-iid}
\end{figure}

To evaluate the impact of partial participation (PP), we vary the number of clients involved in each communication round, randomly selecting 1, 2, 5, or 10 out of a total of 10 clients. While one might assume that involving more clients per round would result in better performance, our experiments under IID conditions reveal minimal performance differences across varying levels of participation, as shown in Fig.~\ref{fig:pp-iid}.

This result suggests that in the IID setting, an update from a single client remains valid for others, allowing them to resume training locally from that state without significant performance degradation. Although fewer clients participating per round introduce more fluctuations, the final accuracy still converges to nearly the same level.

\vspace{-3mm}

\subsubsection{Impact of Imbalanced Data.} \label{sec:iid-imbalance}
\begin{figure}[h]
\centering
\includegraphics[width=\linewidth]{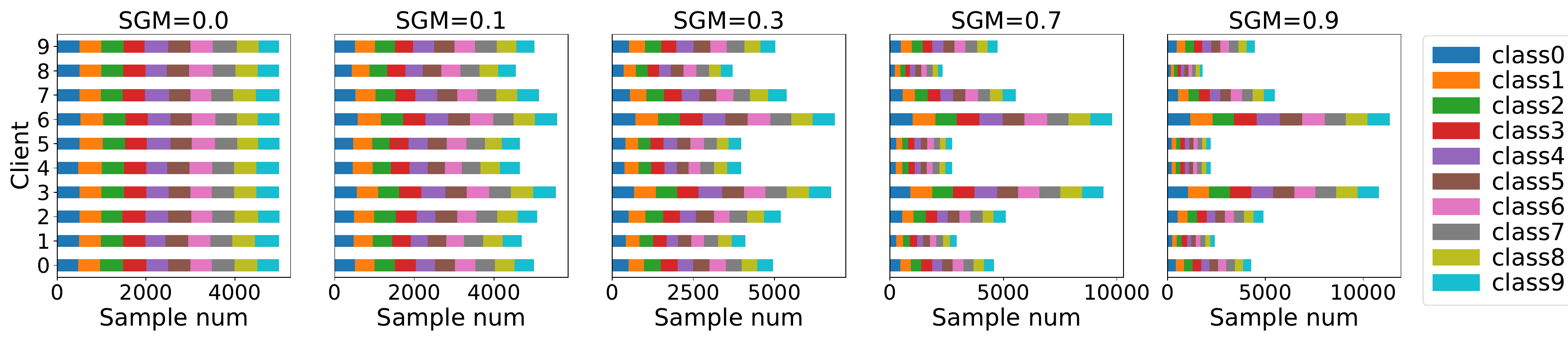}
\vspace{-5mm}
\caption{Data partitioning with IID labels across different levels of SGM} 
\label{fig:iid}
\end{figure}

In our experiments with imbalanced datasets, we construct the environment using a Standard Gaussian Mixture (SGM) model. Here, an SGM value of 0 corresponds to a balanced dataset, while higher SGM values indicate increasing levels of data imbalance, as illustrated in Fig.~\ref{fig:iid}.

\vspace{-5mm}

For imbalanced datasets, we explore two aggregation strategies: \textbf{weighted aggregation}, used in FedAvg, where the contribution of each client is proportional to its dataset size, and \textbf{naive aggregation}, where updates from all clients are averaged equally, regardless of dataset size.

\vspace{-3mm}
\begin{figure}[h]
\centering
\includegraphics[width=\linewidth]{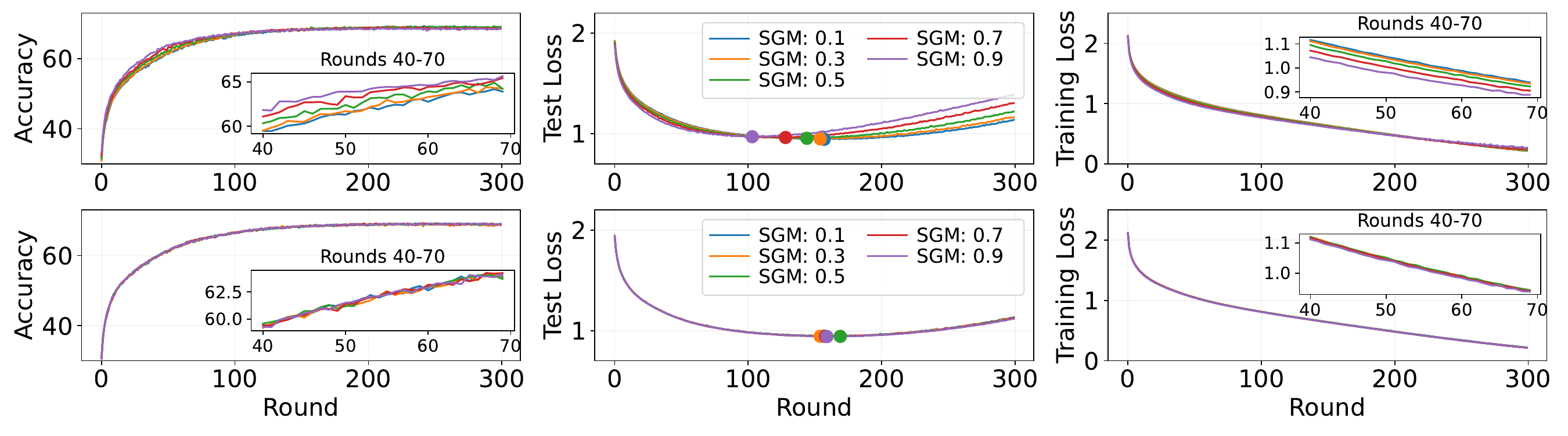}
\vspace{-3mm}
\caption{Performance under varying levels of data imbalance. The first row shows results using weighted aggregation, while the second row uses naive averaging.} 
\label{fig:weighted-vs-naive}
\end{figure}
\vspace{-3mm}

Contrary to intuitive expectations that increased imbalance would degrade performance, our experiments reveal that \textbf{weighted aggregation} actually accelerates convergence in IID settings. As shown in the first row of Fig.~\ref{fig:weighted-vs-naive}, models converge faster as imbalance increases, especially for SGM 0.9 (purple), which shows higher accuracy, earlier overfitting, and lower training loss between rounds 40-70. This is because, in weighted aggregation, clients with larger datasets exert more influence, effectively increasing the total number of updates. For instance, when Client A has 10 data points and Client B has 90, the weighted average of updates is $ \frac{10 \times 10 + 90 \times 90}{100} = 82 $. In contrast, with balanced datasets (50 data points each), the average number of updates is $ \frac{50 \times 50 + 50 \times 50}{100} = 50 $. Thus, in imbalanced scenarios, the total number of updates per round is higher, leading to faster convergence. Furthermore, when data is IID, the imbalance does not negatively impact performance—in fact, it can be advantageous.

We also conducted experiments using \textbf{naive aggregation}, which showed minimal performance differences across varying levels of imbalance, as shown in the second row of Fig.~\ref{fig:weighted-vs-naive}. This outcome can be understood by considering the number of updates: whether the dataset sizes are imbalanced (10 and 90) or balanced (50 and 50), naive averaging still results in an equal average of 50 updates per round. Interpreting updates as vectors, we can infer that when data is IID, the update vectors are aligned in direction, meaning that as long as the magnitudes are adjusted appropriately, the combined update remains the same. This suggests that in IID settings, updates from individual clients are consistently aligned in direction.

\vspace{-3mm}

\subsection{Training Dynamics Under Non-IID Setting}

\subsubsection{Non-IID with Dirichlet Distribution}
\begin{figure}[h] \centering \includegraphics[width=\linewidth]{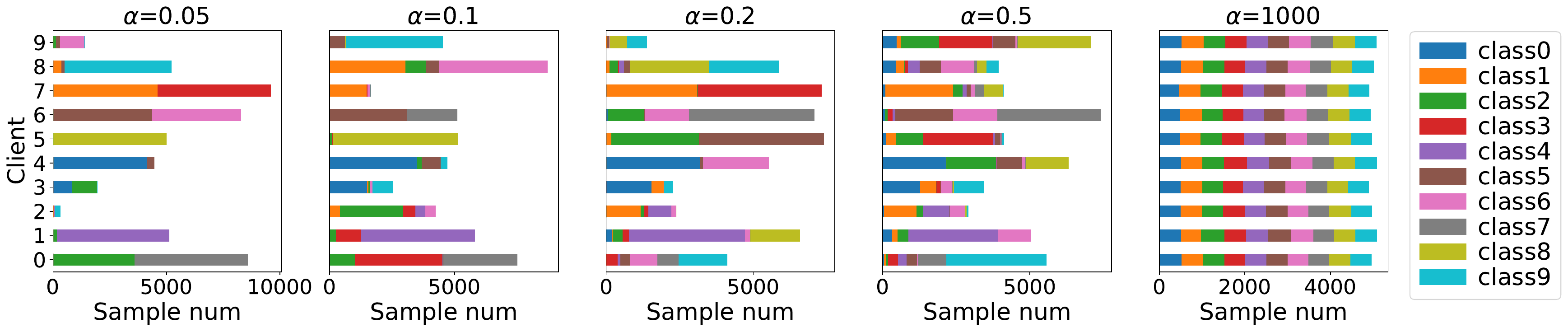} 
\vspace{-3mm}
\caption{Data partitioning with varying Dirichlet parameter \(\alpha\).}
\label{fig:dirichlet} \end{figure}

Next, we experiment non-IID label distribution with a Dirichlet distribution for data partitioning across clients, which introduces varying degrees of non-IID behavior. In this setup, the degree of non-IID distribution is controlled by the Dirichlet $\alpha$: a lower $\alpha$ leads to more distinct distributions across clients, while a higher $\alpha$ results in more similar distributions. The resulting distributions are visualized in Fig.~\ref{fig:dirichlet}.

\vspace{-3mm}
\vspace{-3mm}

\begin{figure}[h]
\centering
\includegraphics[width=\linewidth]{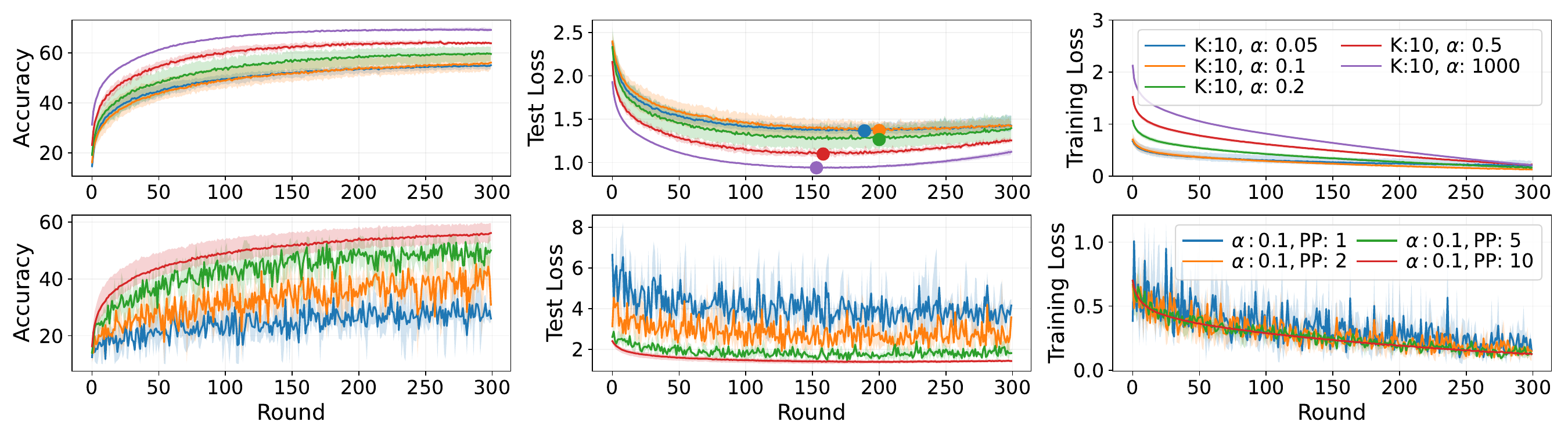}
\caption{Performance across varying $\alpha$ (Top) and PP (Bottom).}

\label{fig:non-iid-result}
\end{figure}

In Fig.~\ref{fig:non-iid-result}, each line represents the average of three independent runs, with the shaded region indicating the min-max range across these runs. This approach accounts for the randomness introduced by the non-IID distribution of the dataset. Additionally, in the case of PP, the selection of clients in each round can have a significant impact on the results.

We observe that as the data distribution becomes more non-IID (i.e., as $\alpha$ decreases), model performance degrades (from 69\% to 55\%) and convergence slows. While lower $\alpha$ values do eventually converge within the limited communication rounds, the resulting solutions are highly suboptimal. Furthermore, unlike in IID settings, the level of PP plays a critical role in FL performance. When $\alpha$ is fixed at 0.1, lower levels of PP lead to significantly lower final accuracy, higher test loss, and increased fluctuations compared to full participation. This indicates that updates from different clients are not valid for others.

\subsubsection{Impact of Hyperparameters under Non-IID Condition}

\begin{figure}
\includegraphics[width=\linewidth]{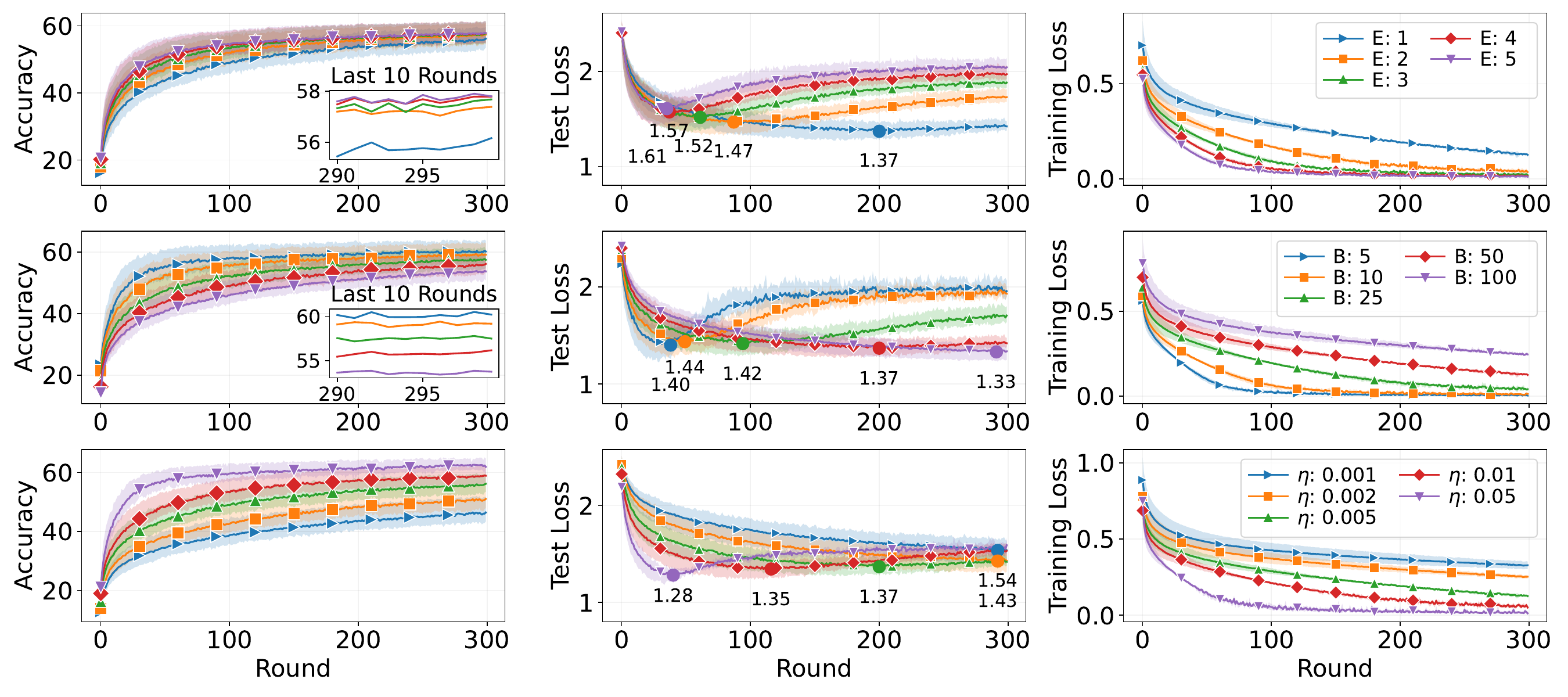}
\vspace{-5mm}
\caption{Test accuracy, test loss, and training loss for $K=10$. The first row shows results for varying $E$, the second for different $B$, and the third for various $\eta$.}

\label{fig:non-iid-diff-param}
\end{figure}

We investigate the effects of varying hyperparameters, as shown in Fig.~\ref{fig:non-iid-diff-param}.

\vspace{-5mm}

First, increasing $E$ results in faster convergence, and unlike the IID setting, a higher number of epochs leads to improved final performance. The difference in accuracy between the smallest and largest $E$ is $+$1.7\% (from 56.16\% to 57.86\%), whereas in the IID setting, this difference is $-$1.55\% (from 68.75\% to 67.2\%). Second, lower values of $B$ result in both higher final accuracy and faster convergence. The accuracy gap between the smallest and largest $B$ is $+$6.45\% (from 53.74\% to 60.19\%), compared to $+$3.57\% (from 67.77\% to 71.34\%) in the IID case. Finally, $\eta$ plays a crucial role in the non-IID scenario. The accuracy difference between the smallest and largest $\eta$ is $+$15.48\% (from 46.43\% to 61.91\%), compared to $+$7.79\% (from 63.22\% to 71.01\%) in the IID setting. Notably, lowest test loss decreases as $\eta$ increases.

\vspace{-3mm}
\section{Understanding FL from a Loss Landscape Perspective}\label{sec:how-it-works}
\vspace{-3mm}

We have performed a series of experiments across datasets ranging from IID to non-IID, incorporating both partial participation and imbalanced data. As expected, the performance of the global model declines as the level of non-IIDness increases. The most commonly cited explanation for this behavior is client drift. \textit{But what drives this phenomenon?}

To uncover the underlying causes, we examine the problem through the lens of the loss landscape~\cite{li2018visualizing}. The loss landscape is shaped by the interaction between the loss function, dataset, and model parameters. It provides valuable insights into how the loss changes as model parameters are updated for a fixed dataset.

\begin{figure}
\centering
\includegraphics[width=0.7\textwidth]{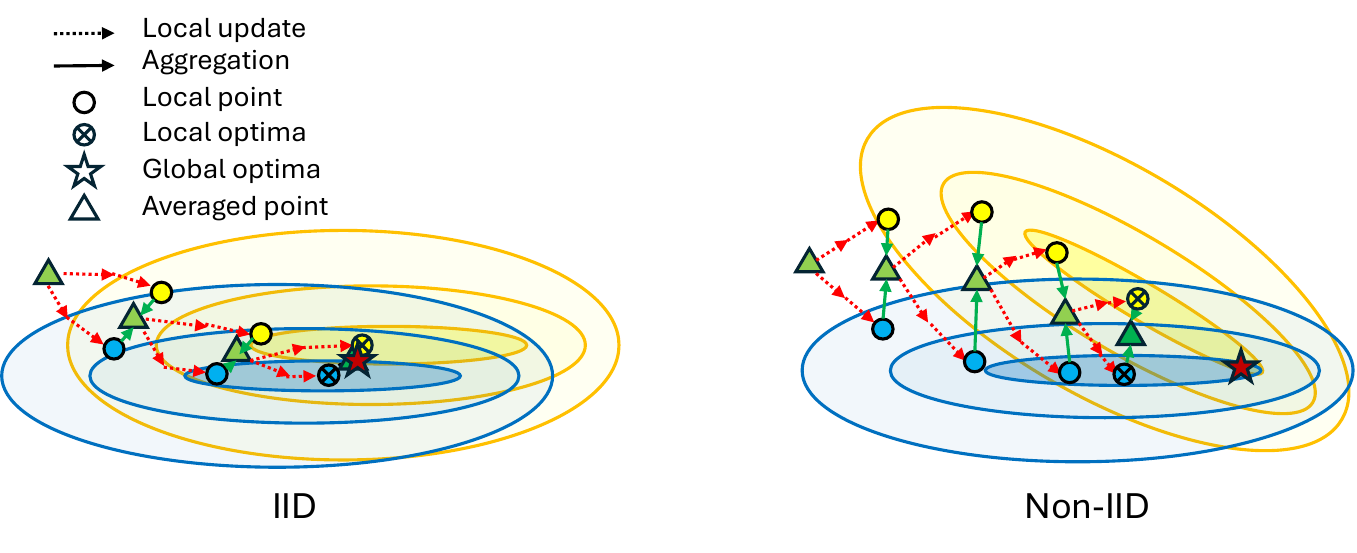}
\caption{Toy example illustrating the loss landscape and update paths for two clients with IID and Non-IID datasets.} \label{fig:loss-landscape}
\end{figure}

In FL, each client operates on a distinct dataset, meaning each experiences a unique loss landscape~\cite{al-shedivat2021federated,zhang2022federated}. Even though all clients use the same model and loss function, the variations in datasets lead to different optimization paths. As a result, updates from different clients are often misaligned, causing client drift. As each client optimizes locally, their models may converge to different local optima. When these locally optimized models are aggregated, the result may fail to reach the global optimum, as illustrated in Fig.~\ref{fig:loss-landscape}. 

In contrast, under IID conditions, these differences are minimized as the clients' loss landscapes are more similar, leading to better-aligned update directions. Since directly visualizing the loss landscape in FL is difficult due to the high dimensionality and variability in client data, we instead performed two additional experiments to substantiate this observation.

\begin{figure}
\centering
\includegraphics[width=\linewidth]{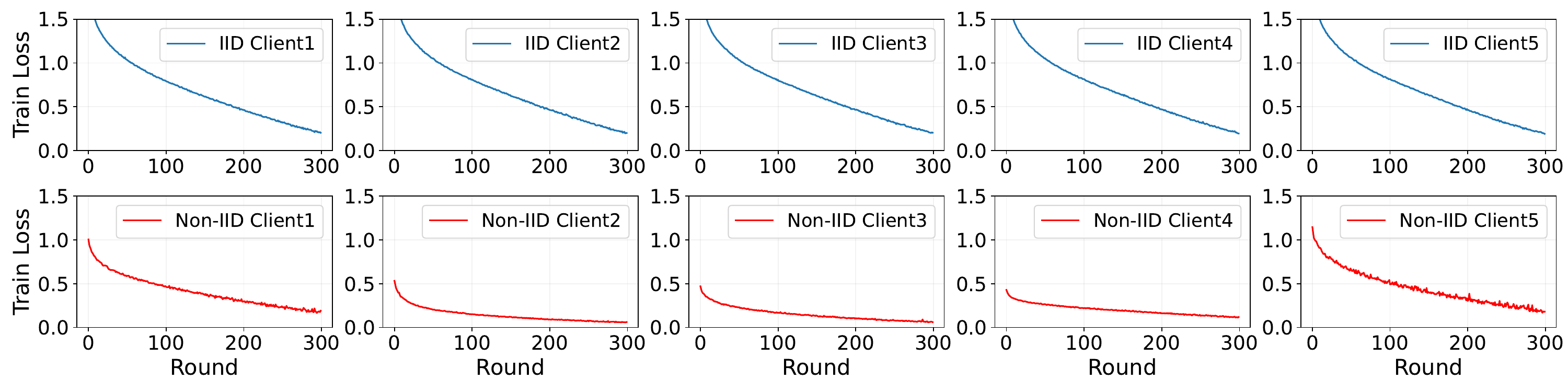}

\caption{Client training loss: Top row - IID, Bottom row - Non-IID.}
\label{fig:individual-loss}
\end{figure}

First, we selected 5 clients from both IID and non-IID settings when $K$ is 10 and traced their individual training loss. As shown in Fig.~\ref{fig:individual-loss}, under non-IID conditions, individual loss values decrease at different rates, while in the IID setting, the loss values for each client decrease more uniformly. This suggests that the local optimization processes in non-IID settings follow different loss landscapes, leading to varying loss values across clients.

\begin{figure}
\centering
\includegraphics[width=\linewidth]{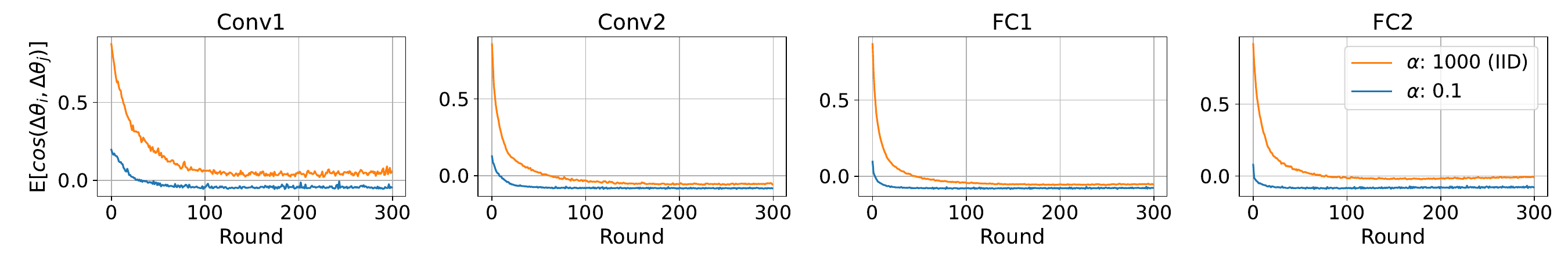}
\vspace{-5mm}
\caption{Cosine similarity of local updates from different clients for each layer} 
\label{fig:update-alignment}
\end{figure}

Secondly, in Fig.~\ref{fig:update-alignment}, we measure the layer-wise (conv1, conv2, fc1, and fc2) cosine similarity of local updates ($\Delta\theta$) between pairs of clients, averaged across all combinations from 10 clients. As anticipated, in the IID setting, the cosine similarity between client updates remains consistently higher compared to the non-IID setting.

\vspace{-5mm}
These findings show that non-IID conditions lead to more diverse loss landscapes and less aligned client updates, while IID settings promote better alignment through consistent loss landscapes.

To guide the global model toward optimal performance in non-IID settings, two main strategies emerge:
\vspace{-2mm}
\begin{enumerate}
    \item \textbf{Adjusting the update path}, where optimization steps are tailored to direct the model toward the global optimum without altering the underlying loss landscape.
    \item \textbf{Modifying the loss landscape}, which aims to increase the overlap between local optima of individual clients, thus promoting more consistent convergence across clients.
\end{enumerate}
\vspace{-3mm}
For example, methods like SCAFFOLD~\cite{karimireddy2020scaffold}, which employs control variates, and FedOpt~\cite{reddi2021adaptive}, which uses server-side optimizers (e.g., Adam, Yogi) or weight perturbation~\cite{qu2022generalized,lee2024fedsol}, focus on refining update directions. In contrast, strategies that modify the loss landscape add terms to the objective function~\cite{li2020federated,li2021model,lee2022preservation}. Hybrid approaches like FedDyn~\cite{durmus2021federated} address both aspects simultaneously. However, a deeper analysis of these methods is necessary to fully understand their impact.

\vspace{-4mm}
\section{Discussion and Conclusion}
\vspace{-3mm}

Our experiments consistently show that a higher $\eta$ and lower $B$ improve performance in both IID and non-IID settings. According to our categorization, these hyperparameters are part of \textbf{adjusting the update path}. A larger $\eta$ leads to bigger steps in gradient descent, which can skip sharp minima and converge on flatter ones, although too large of a step risks divergence. Similarly, a smaller $B$ aids in escaping sharp minima and encourages convergence to flatter minima~\cite{li2018visualizing}, improving generalization~\cite{keskar2017on} and potentially increasing the likelihood of finding overlapping local optima, a topic that requires further investigation.

In conclusion, our exploration of optimization algorithms, from GD to FL, reveals that client drift stems from inconsistencies in loss landscapes. Based on these findings, we suggest that methods addressing performance degradation in non-IID settings can be categorized into two main strategies: adjusting the update path or modifying the loss landscape. While the motivations behind these methods may vary, they generally align with these categories. We hope this analysis provides valuable insights for new FL researchers and contributes to ongoing discussions in the field.

\begin{credits}
\vspace{-3mm}
\subsubsection{\ackname} The authors acknowledge the Research Council of Norway and the industry partners of NCS2030 – RCN project number 331644 – for their support.

\vspace{-3mm}
\subsubsection{\discintname}
The authors have no competing interests to declare that are relevant to the content of this article.
\end{credits}
\vspace{-4mm}
%
%
%
\bibliographystyle{splncs04}
\bibliography{paper}

\end{document}